\documentclass{article}
\usepackage{spconf,amsmath,graphicx}
\usepackage{booktabs}
\usepackage{subcaption}

\usepackage{enumitem}
\setlist{nosep, leftmargin=14pt}

\usepackage{mwe}

\title{FORESEE-MS: Toward Prospective Spatial Forecasting of Future Multiple Sclerosis Lesions from Baseline FLAIR MRI}

\name{Maryam Akhavan Aghdam, Sudip Mittal, Noorbakhsh Amiri Golilarz, Shahram Rahimi} 
\address{Department of Computer Science, The University of Alabama,
Tuscaloosa, AL, USA}

\begin{document}

\raggedbottom
\setlength{\textfloatsep}{6pt plus 2pt minus 2pt}
\setlength{\intextsep}{6pt plus 2pt minus 2pt}

\maketitle

\begin{abstract}
Traditional multiple sclerosis (MS) image analysis has primarily focused on retrospectively detecting and segmenting lesions. This study investigates whether baseline FLAIR MRI can identify regions at risk for future lesion formation beyond population-level spatial patterns. We introduce \textit{FORESEE-MS}, an evaluation protocol that uses a fold-specific population-prior atlas as a baseline and assesses predictions using pooled voxel-level discrimination (AUROC and AUPRC) and within-patient lesion localization (Recall@Top-$K$\%), with patient-level bootstrap confidence intervals. We evaluated five deep learning models (3D U-Net, Attn+\allowbreak ResBlock+\allowbreak VAE, ViT, and two SwinUNETR variants) using 5-fold cross-validation on 39 subjects from the MSSEG-2 dataset. The atlas achieved the highest pooled voxel-level AUROC and AUPRC, with no deep learning model reliably outperforming it on either metric. However, the 3D U-Net and Attn+\allowbreak ResBlock+\allowbreak VAE achieved higher mean voxel-level within-patient Recall@Top-$K$\% than the atlas at all evaluated thresholds, although most paired confidence intervals for the differences included zero. These findings suggest that baseline FLAIR may contain patient-specific predictive information, but in this small cohort, it remains uncertain whether this information improves lesion forecasting beyond a population-prior atlas.
\end{abstract}


\vspace{-10pt}
\section{Introduction}
\label{sec:intro}
\vspace{-10pt}

Multiple sclerosis (MS) is a chronic demyelinating disease of the central nervous system, characterized by focal white matter lesions on magnetic resonance imaging (MRI)~\cite{reich2018multiple}. Prior MS image-analysis work has largely addressed \emph{retrospective} lesion segmentation~\cite{commowick2021msseg2,carass2017longitudinal}: given one or more scans, identify pathology that is already present. This leaves open a different question: given only a patient's baseline scan, can we anticipate \emph{where} new pathology is likely to emerge?

Prior evidence suggests this may be possible. Normal-appearing white matter (NAWM) that later becomes a lesion has been reported to show elevated T2 signal and reduced magnetization transfer ratio before lesion formation~\cite{nawm_prelesional}, and fluid-attenuated inversion recovery (FLAIR) is closely related to T2-weighted contrast. We therefore ask: \emph{can baseline FLAIR MRI identify regions at risk of future lesion formation, beyond population-level spatial patterns?}

Existing work on future lesion activity mostly operates at the patient or region level. Baseline MRI has been used to predict future new and enlarging T2 lesion counts~\cite{durso2022personalized}, and a treatment-aware diffusion model generates voxel-level future lesion masks from MRI and treatment information~\cite{favero2025diffusion}. A recent study extracted radiomic features from prelesional tissue in prior scans and trained classifiers to distinguish them from those of sampled control normal-appearing white matter (NAWM)~\cite{kelly2024radiomic}. In contrast, we evaluate dense voxel-wise risk maps across the entire brain using baseline FLAIR MRI and derived spatial information, without incorporating treatment or clinical data.

We make two primary contributions. First, we introduce \textit{FORESEE-MS}, an evaluation protocol for forecasting future MS lesion locations from baseline MRI. We construct a fold-specific population-prior atlas as a baseline and evaluate predictions using two complementary approaches: pooled voxel-level discrimination, measured by the area under the receiver operating characteristic curve (AUROC) and the area under the precision-recall curve (AUPRC), and within-patient lesion localization, measured by Recall@Top-$K$\%, with patient-level bootstrap confidence intervals (CIs). Second, we evaluate five deep learning models, including a 3D U-Net, an Attn+\allowbreak ResBlock+\allowbreak VAE, a vision transformer (ViT), and two SwinUNETR variants, using this protocol on the MSSEG-2 dataset~\cite{commowick2021msseg2}.

\vspace{-10pt}

\section{Method}
\label{sec:method}
\vspace{-10pt}

\begin{figure*}[t]
\centering
\includegraphics[width=0.9\linewidth]{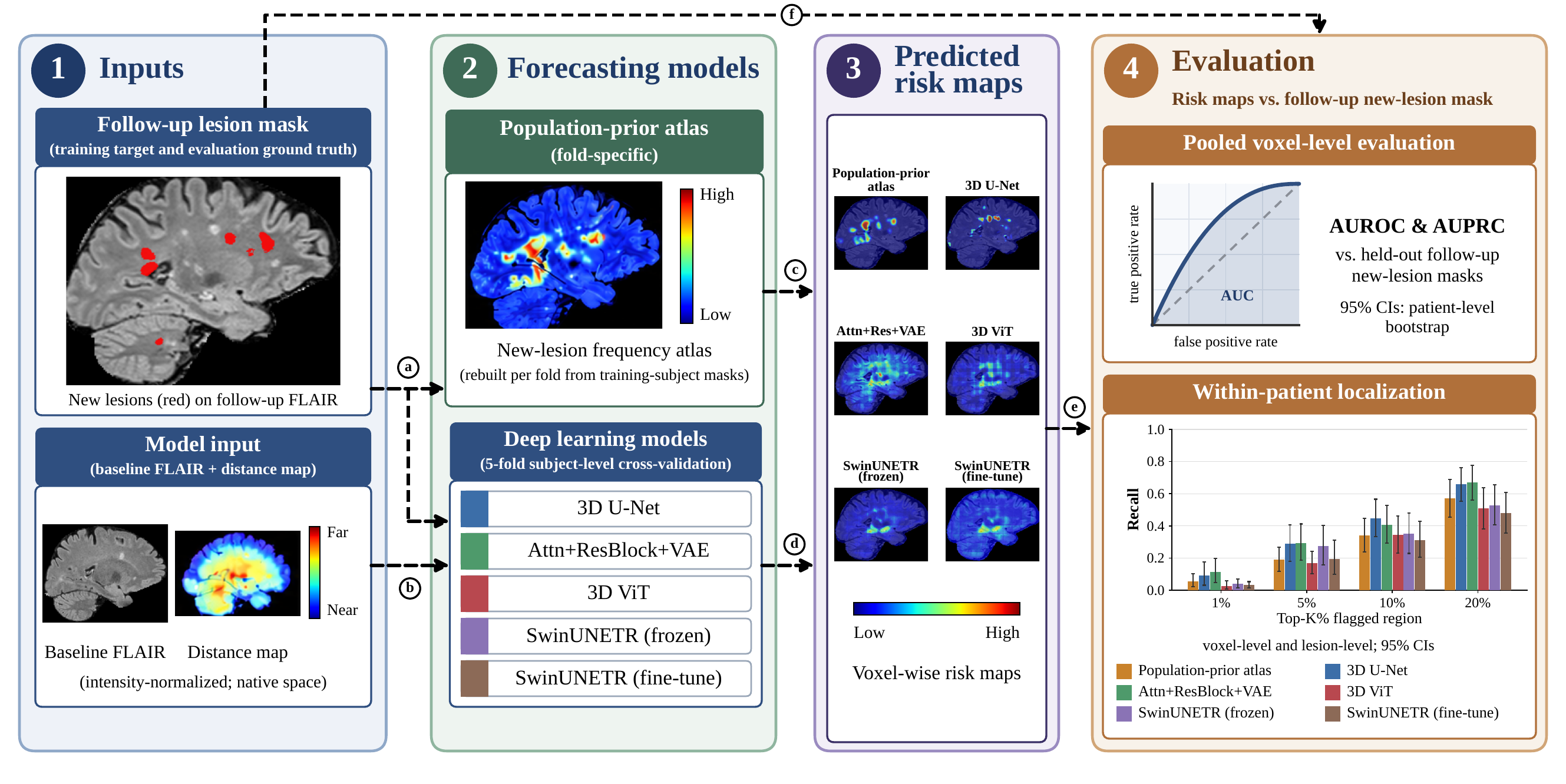}
\vspace{-0.4cm}
\caption{Overview of the \textit{FORESEE-MS} workflow. (a) Training-subject follow-up new-lesion masks are used to construct the fold-specific population-prior atlas and supervise the deep learning models. (b) At inference, the models receive only baseline FLAIR and the distance-to-brain-boundary map. (c,d) The atlas and each deep learning model independently produce voxel-wise risk maps for future lesion formation. (e) Risk maps are evaluated using pooled voxel-level AUROC and AUPRC and within-patient Recall@Top-$K$\%. (f) Held-out follow-up lesion masks are used only for evaluation.}
\label{fig:pipeline}
\end{figure*}

\textbf{Problem setting.} Fig.~\ref{fig:pipeline} presents an overview of the FORESEE-MS workflow. For each subject, the forecasting methods produce voxel-wise risk maps for new-lesion formation at follow-up. Each deep learning model receives two input channels derived solely from the baseline scan: intensity-normalized FLAIR and a Euclidean distance-to-brain-boundary map. Follow-up new-lesion masks serve as training targets and evaluation ground truth, while training-fold masks also contribute to the population-prior atlas. We used held-out follow-up scans and lesion masks only for evaluation, not for training, model selection, or risk-map generation.

\textbf{Population-prior baseline.} To obtain a population-level baseline, we built a voxel-wise atlas of new-lesion frequency. We registered each training subject's FLAIR to the ICBM152 2009a template~\cite{fonov2011unbiased} using affine and SyN registration~\cite{avants2008syn} (ANTsPy), and applied the same transform to that subject's new-lesion mask. Each template voxel was assigned the fraction of training subjects with a new lesion there (among subjects whose brain mask covers it), followed by Gaussian smoothing ($\sigma=2$\,mm). We rebuilt the atlas for each cross-validation fold (Section~\ref{ssec:setup}) from the training subjects of that fold only. We then warped it into each held-out subject's native space using the inverse of that subject's template registration. This approach ensured that held-out subjects contributed only their anatomical information, not their lesion data.

\textbf{Models.} We compared five deep learning models; the loss and training details are in Section~\ref{ssec:setup}. (1) A shallow 3D U-Net~\cite{cicek2016unet3d} ($\sim$1.4M parameters). (2)  An attention- and residual-enhanced 3D U-Net (Attn+ResBlock+VAE) incorporating attention gates~\cite{oktay2018attention}, pre-activation residual blocks, and a VAE-based regularization branch that reconstructs the input FLAIR~\cite{myronenko2018autoencoder} ($\sim$2.4M parameters). (3) A lightweight 3D ViT~\cite{dosovitskiy2021vit} with a patch-embedding layer, a 6-layer transformer encoder, and a convolutional upsampling decoder producing voxel-wise predictions ($\sim$3.5M parameters). Models (1)--(3) were trained from scratch. (4--5) SwinUNETR~\cite{tang2022swin} ($\sim$62M parameters), initialized from BrainSegFounder~\cite{cox2024brainsegfounder}, a brain MRI foundation model pretrained with self-supervision on UK Biobank brain MRI and then fine-tuned for brain tumor segmentation. We tested SwinUNETR with two transfer strategies: in the frozen variant, we keep the Swin transformer encoder's pretrained weights fixed and train the remaining layers; in the fine-tuned variant, we train all layers on our data. For both variants, we re-initialized three input-facing layers (the patch-embedding projection and two convolutions in the first encoder block) to accommodate our two-channel input, and the output layer for our single-channel output.

\vspace{-10pt}
\section{Experimental Results and Discussion}
\label{sec:results}
\vspace{-8pt}
\subsection{Experimental Setup}
\label{ssec:setup}
\vspace{-5pt}

\textbf{Data and preprocessing.} We used the MSSEG-2 dataset~\cite{commowick2021msseg2}, which includes 40 training subjects with baseline and follow-up FLAIR images and consensus new-lesion masks. The 60-subject test partition lacks public ground-truth labels; therefore, we used only the 40 labeled training subjects. We preprocessed the volumes using the official MSSEG-2 pipeline~\cite{commowick2021msseg2}, which performs brain extraction and bias-field correction, and then applied z-score normalization within the brain mask. During quality control, we excluded one subject whose lesion mask was largely outside the brain mask (27.6\% overlap compared to 98--100\% for all others), resulting in 39 subjects. New lesions comprise approximately 0.035\% of brain voxels (58{,}292 of $1.65 \times 10^8$).

\textbf{Training.} We used 5-fold subject-level cross-validation and trained models on $64^3$ patches. For lesion-positive subjects, approximately half of the patches were centered on new-lesion voxels with random offsets. Data augmentation included random axis flips and additive Gaussian noise on the FLAIR channel. Whole-volume inference used overlapping sliding windows (stride 32), with predictions averaged across overlapping regions. All models minimized a segmentation loss, $\mathcal{L}_{\mathrm{seg}}=\mathcal{L}_{\mathrm{BCE}}+\mathcal{L}_{\mathrm{Tversky}}$, with a positive-class weight of 50 for binary cross-entropy (BCE) and Tversky parameters $\alpha=0.3$ and $\beta=0.7$~\cite{salehi2017tversky}. Additionally, Attn+ResBlock+VAE used the combined loss $\mathcal{L}=\mathcal{L}_{\mathrm{seg}}+0.1(\mathcal{L}_{\mathrm{MSE}}+0.1\mathcal{L}_{\mathrm{KL}})$, where MSE measures input FLAIR reconstruction error and KL regularizes the VAE latent representation.

The U-Net, Attn+\allowbreak ResBlock+\allowbreak VAE, and ViT were trained for 120 epochs (batch size~4), while both SwinUNETR variants were trained for 60 epochs (batch size~2). We used the Adam optimizer (weight decay $10^{-5}$) for the U-Net and Attn+ResBlock+VAE, and AdamW (weight decay $10^{-4}$) for ViT and SwinUNETR. Initial learning rates were $3\times10^{-4}$ for the U-Net, Attn+ResBlock+VAE, and ViT; $10^{-4}$ for frozen SwinUNETR; and $2\times10^{-5}$ for fine-tuned SwinUNETR.

\textbf{Evaluation metrics.}
We evaluated predicted risk maps against follow-up new-lesion masks using pooled voxel-level AUROC and AUPRC across all out-of-fold brain voxels to assess discrimination between future lesion and non-lesion voxels. We assessed within-patient lesion localization using Recall@Top-$K$\% ($K\in\{1,5,10,20\}$), which measures the fraction of future lesion voxels or lesion objects captured within each subject's top-$K$\% highest-risk brain voxels. We averaged voxel-level recall across 28 lesion-positive subjects, whereas lesion-level recall was computed across 147 new-lesion objects, counting an object as detected if any of its voxels overlapped the selected region. Atlas ties were handled using exact expected recall under uniform random tie-breaking. Because baseline lesion masks were unavailable, we considered all brain voxels eligible for selection, including those containing pre-existing lesions.

We estimated 95\% CIs using patient-level bootstrap resampling, with 1{,}000 resamples for pooled AUROC and AUPRC and 10{,}000 for Recall@Top-$K$\%. All voxels and lesion objects from each subject were resampled together. Paired comparisons used the same resamples across methods. For each model and metric $M$, we computed $\Delta=M_{\mathrm{model}}-M_{\mathrm{atlas}}$ and reported the mean bootstrap difference with a 95\% percentile CI. Negative differences favored the atlas. The 40 recall comparisons and 10 pooled AUROC and AUPRC comparisons were not adjusted for multiple testing; thus, intervals excluding zero were interpreted as nominal, exploratory findings.

\vspace{-10pt}
\subsection{Quantitative Results}
\label{ssec:quantitative}

\textbf{Pooled voxel-level discrimination.} Table~\ref{tab:main} summarizes pooled AUROC and AUPRC. The population-prior atlas achieved the highest AUROC (0.747) and AUPRC ($1.24\times10^{-3}$), while all five deep learning models had lower AUROC point estimates (0.643--0.682). All paired AUROC CIs included zero. For the U-Net, the mean paired bootstrap difference ($\Delta$) was $-0.043$ (95\% CI [$-0.237$, $+0.120$]) for AUROC and $-0.0004$ (95\% CI [$-0.0015$, $+0.0005$]) for AUPRC. Two nominal paired AUPRC comparisons favored the atlas over the lightweight 3D ViT ($\Delta=-0.00062$, 95\% CI [$-0.00163$, $-0.00002$]) and fine-tuned SwinUNETR ($\Delta=-0.00071$, 95\% CI [$-0.00169$, $-0.00001$]). These results suggest that population-level spatial patterns provide useful information for forecasting future lesion locations, while baseline FLAIR-based deep learning models did not demonstrate a reliable improvement in pooled discrimination over the population-prior atlas.

\begin{table}[htb]
\centering
\caption{Pooled voxel-level AUROC/AUPRC from 5-fold cross-validation across 39 subjects, with 95\% CIs estimated from 1{,}000 patient-level bootstrap resamples. Best values in each column are shown in bold.}
\label{tab:main}
\footnotesize
\setlength{\tabcolsep}{3pt}
\begin{tabular}{lcc}
\toprule
Method & AUROC [95\% CI] & AUPRC ($\times10^{-3}$) [95\% CI] \\
\midrule
Attn+ResBlock+VAE & 0.643 [0.465, 0.867] & 0.82 [0.28, 2.54] \\
3D U-Net & 0.680 [0.503, 0.866] & 0.78 [0.30, 2.01] \\
SwinUNETR (frozen) & 0.682 [0.645, 0.755] & 0.54 [0.25, 0.98] \\
SwinUNETR (fine-tune) & 0.677 [0.619, 0.759] & 0.54 [0.24, 0.98] \\
Lightweight 3D ViT & 0.681 [0.548, 0.814] & 0.61 [0.21, 1.33] \\
\textbf{Population atlas} & \textbf{0.747} [0.682, 0.777] & \textbf{1.24} [0.31, 2.63] \\
\bottomrule
\end{tabular}
\end{table}

\begin{figure}[t]
\centering
\includegraphics[width=0.9\linewidth]{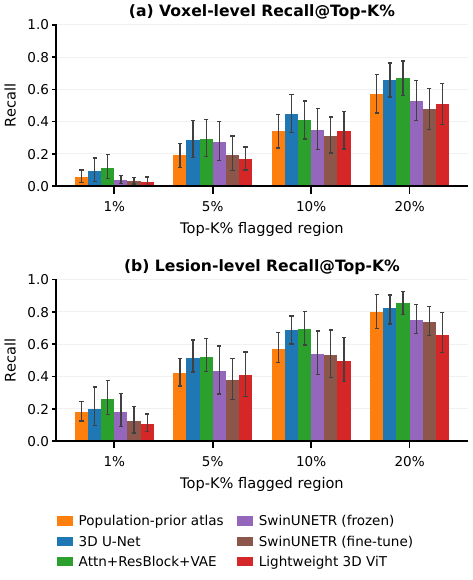}
\vspace{-0.45cm}
\caption{Recall@Top-$K$\% across methods: (a) voxel-level recall averaged over 28 lesion-positive subjects and (b) lesion-level recall pooled over 147 lesions. Error bars indicate 95\% patient-level bootstrap CIs for each method.}
\label{fig:recall}
\end{figure}

\textbf{Within-patient localization.} Fig.~\ref{fig:recall} shows Recall@Top-$K$\% across the 28 lesion-positive subjects. At $K=20\%$, mean voxel-level recall was 0.671 for Attn+ResBlock+VAE and 0.660 for the U-Net, compared with 0.572 for the atlas; the ViT and both SwinUNETR variants reached 0.481--0.529. Both the U-Net and Attn+ResBlock+VAE also achieved higher mean recall than the atlas at $K=1\%$, $5\%$, and $10\%$. However, uncertainty remained substantial: among the 40 paired recall comparisons, six 95\% CIs excluded zero (four favoring the U-Net or Attn+ResBlock+VAE over the atlas and two favoring the atlas over the ViT), while the remaining 34 included zero. These results suggest that baseline FLAIR may contain patient-specific information useful for localizing future lesions, but it remains uncertain whether this improves localization beyond the population-prior atlas in this small MSSEG-2 cohort.

\vspace{-10pt}
\subsection{Qualitative Results}
\label{ssec:qualitative}

\begin{figure}[!t]
\centering
\begin{subfigure}{\linewidth}
\centering
\includegraphics[width=0.9\linewidth]{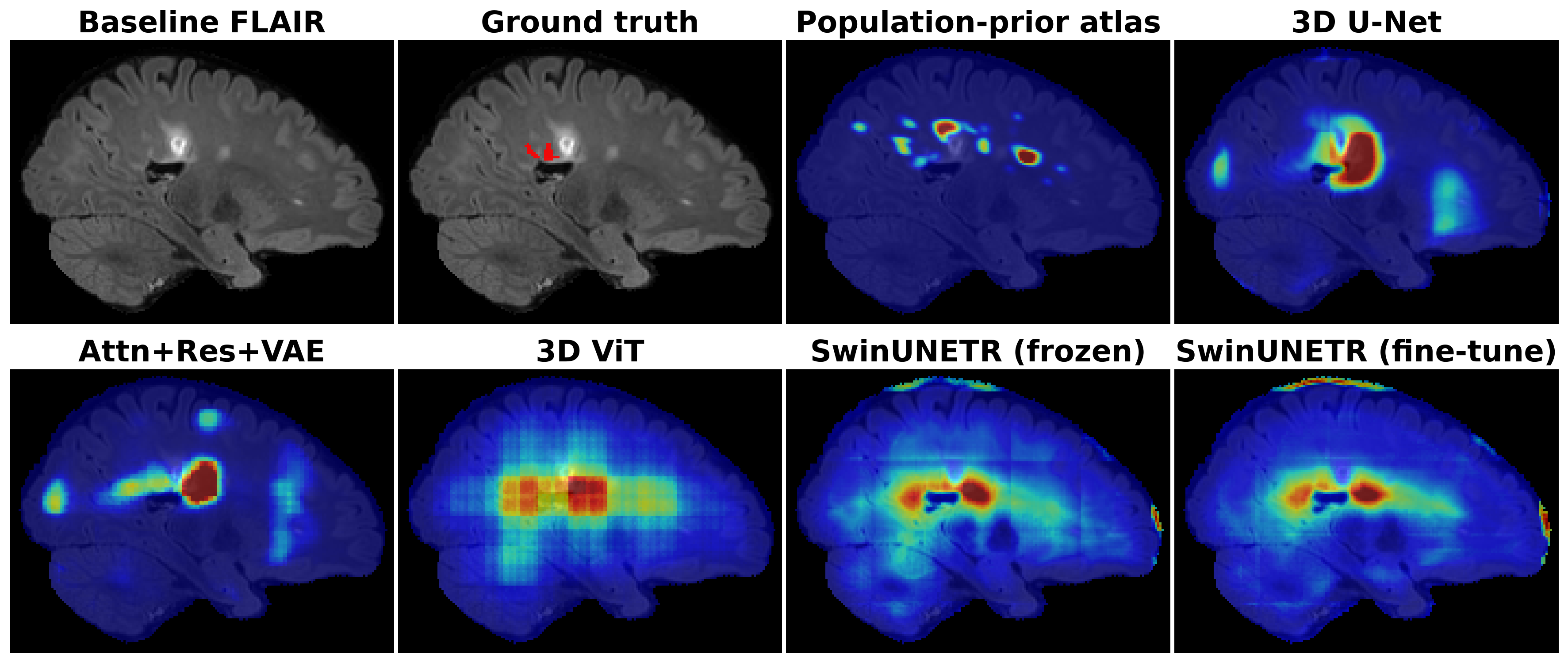}
\caption{Subject 057: Hit (successful localization)}
\end{subfigure}
\begin{subfigure}{\linewidth}
\centering
\includegraphics[width=0.9\linewidth]{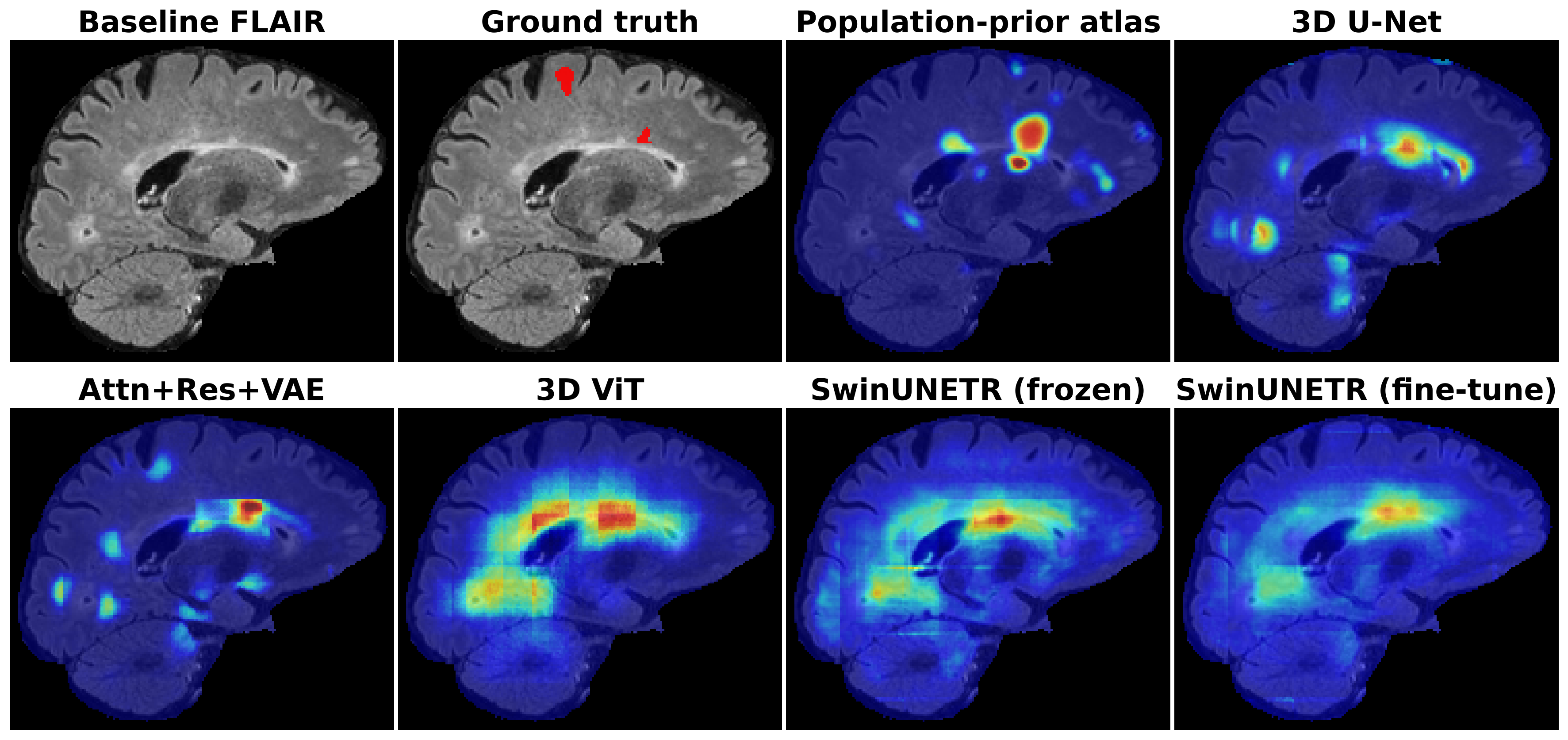}
\caption{Subject 021: Partial hit (partial localization)}
\end{subfigure}
\begin{subfigure}{\linewidth}
\centering
\includegraphics[width=0.9\linewidth]{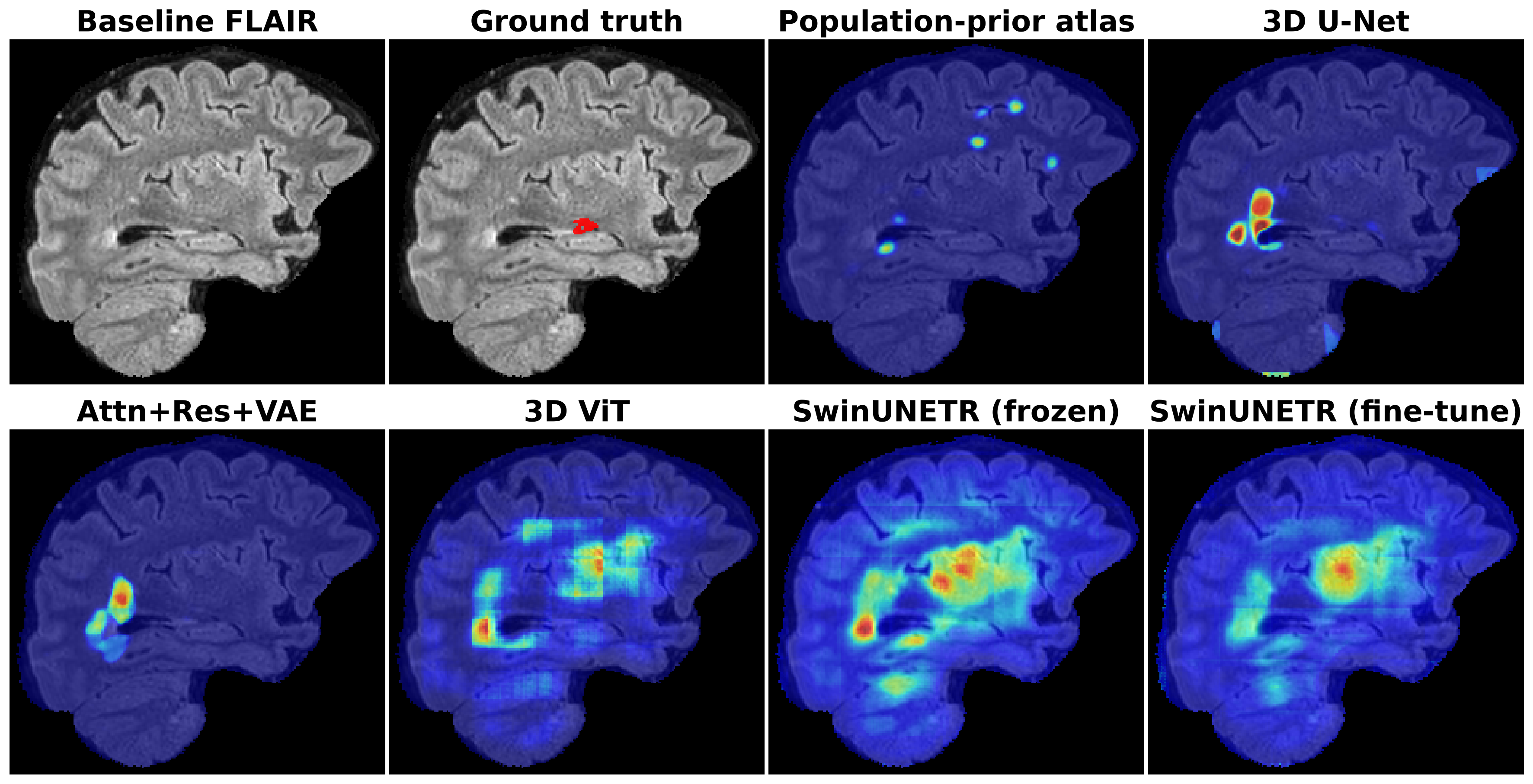}
\caption{Subject 094: Miss (unsuccessful localization)}
\end{subfigure}
\vspace{-0.5cm}
\caption{Qualitative comparison of predicted risk maps for three MSSEG-2 subjects. (a) Subject 057 (hit): deep learning models assign high risk around a ring-shaped FLAIR hyperintensity adjacent to new lesions, while the atlas produces smaller periventricular hotspots. (b) Subject 021 (partial hit): most methods highlight the smaller lesion but miss the larger lesion above it. (c) Subject 094 (miss): no method places a hotspot directly on the lesion in the displayed slice, although some models assign high risk to nearby regions. In each map, blue to red indicates low to high relative risk within the brain.}
\label{fig:qual}
\end{figure}

Fig.~\ref{fig:qual} illustrates three localization outcomes on the slice containing the most new-lesion voxels for each subject: a hit (057), a partial hit (021), and a miss (094). 
The risk maps indicate relative lesion risk rather than calibrated probabilities. In the hit case, the deep learning models assigned high risk to the new lesions or nearby regions; in the partial-hit case, they highlighted only one of two lesions; and in the miss case, they failed to localize the lesion on the displayed slice.

\vspace{-10pt}

\subsection{Discussion}
\vspace{-6pt}
\label{sec:discussion}

No deep learning model reliably outperformed the population-prior atlas on pooled voxel-level AUROC or AUPRC. The atlas achieved the highest AUROC (0.747 versus 0.643--0.682 for the models), suggesting that population-level spatial patterns provide a strong baseline for forecasting future MS lesion locations. However, the 3D U-Net and Attn+ResBlock+VAE achieved higher mean voxel-level within-patient Recall@Top-$K$\% than the atlas at all evaluated thresholds ($K\in\{1,5,10,20\}$). At $K=20\%$, recall was 0.660 for the U-Net and 0.671 for Attn+\allowbreak ResBlock+\allowbreak VAE, compared with 0.572 for the atlas. Most paired CIs included zero. These results do not establish a reliable improvement in patient-specific lesion localization over the atlas. The ViT and SwinUNETR variants showed no consistent improvement over the atlas.

The higher mean voxel-level within-patient recall of the 3D U-Net and Attn+ResBlock+VAE may partly reflect pre-existing lesions visible on baseline FLAIR, which the atlas does not explicitly model, rather than subtle changes in normal-appearing tissue. The small study cohort (39 subjects from MSSEG-2, including 28 with new lesions) limits statistical precision and contributes to wide CIs. In addition, the atlas was constructed using approximately 31 subjects per fold and could benefit from a larger reference cohort. These findings suggest that baseline FLAIR may contain patient-specific predictive information, but whether this information improves lesion forecasting beyond the population-prior atlas remains uncertain in this small cohort.

\vspace{-10pt}
\section{Conclusion}
\label{sec:conclusion}
\vspace{-8pt}

We introduce \textit{FORESEE-MS}, an evaluation protocol to examine whether baseline FLAIR MRI can forecast future MS lesion locations beyond population-level spatial patterns. We evaluated two complementary aspects of forecasting performance: pooled voxel-level discrimination (AUROC and AUPRC) and within-patient lesion localization (Recall@Top-$K$\%). The population-prior atlas achieved the highest pooled voxel-level AUROC and AUPRC, with no deep learning model reliably outperforming it on either metric. However, the 3D U-Net and Attn+ResBlock+VAE achieved higher mean voxel-level within-patient recall than the atlas at all evaluated thresholds ($K\in\{1,5,10,20\}$), although most paired confidence intervals for the differences included zero. The ViT and SwinUNETR variants showed no consistent improvement over the atlas. These findings suggest that baseline FLAIR may contain patient-specific predictive information, but whether this information improves lesion forecasting beyond the population-prior atlas remains uncertain given the small MSSEG-2 cohort. Future work will investigate baseline lesion masks and atlas-informed models using larger multi-site cohorts.

\vspace{-10pt}
\section{Compliance with Ethical Standards}
\label{sec:ethics}
\vspace{-8pt}
This study used retrospective, de-identified MRI data from the MSSEG-2 dataset, derived from the OFSEP cohort and distributed through the Shanoir platform under a Data Use Agreement. No additional ethical approval was required for this secondary analysis of de-identified data.

\vspace{-10pt}
\bibliographystyle{IEEEbib}
\bibliography{strings,refs}

@inproceedings{commowick2021msseg2,
  author    = {Commowick, Olivier and Cervenansky, Fr{\'e}d{\'e}ric and Cotton, Fran{\c{c}}ois and Dojat, Michel},
  title     = {{MSSEG-2} Challenge Proceedings: Multiple Sclerosis New Lesions Segmentation Challenge Using a Data Management and Processing Infrastructure},
  booktitle = {MICCAI 2021 Challenge},
  year      = {2021}
}

@article{carass2017longitudinal,
  author  = {Carass, Aaron and Roy, Snehashis and Jog, Amod and Cuzzocreo, Jennifer L. and Magrath, Elizabeth and Gherman, Adrian and Button, Julia and Nguyen, James and Prados, Ferran and Sudre, Carole H. and others},
  title   = {Longitudinal Multiple Sclerosis Lesion Segmentation: Resource and Challenge},
  journal = {NeuroImage},
  volume  = {148},
  pages   = {77--102},
  year    = {2017},
  doi     = {10.1016/j.neuroimage.2016.12.064}
}

@article{reich2018multiple,
  author  = {Reich, Daniel S. and Lucchinetti, Claudia F. and Calabresi, Peter A.},
  title   = {Multiple Sclerosis},
  journal = {New England Journal of Medicine},
  volume  = {378},
  pages   = {169--180},
  year    = {2018}
}

@article{nawm_prelesional,
  author  = {Elliott, C. and Momayyezsiahkal, P. and Arnold, D. L. and others},
  title   = {Abnormalities in Normal-Appearing White Matter from Which Multiple Sclerosis Lesions Arise},
  journal = {Brain Communications},
  volume  = {3},
  number  = {3},
  pages   = {fcab176},
  year    = {2021},
  doi     = {10.1093/braincomms/fcab176}
}

@misc{favero2025diffusion,
  author       = {Favero, Gian Mario and Luo, Ge Ya and Fathi, Nima and Szeto, Justin and Arnold, Douglas L. and Nichyporuk, Brennan and Pal, Chris and Arbel, Tal},
  title        = {Spatio-Temporal Conditional Diffusion Models for Forecasting Future Multiple Sclerosis Lesion Masks Conditioned on Treatments},
  year         = {2025},
  howpublished = {arXiv preprint arXiv:2508.07006}
}

@inproceedings{cicek2016unet3d,
  author    = {{\c{C}}i{\c{c}}ek, {\"O}zg{\"u}n and Abdulkadir, Ahmed and Lienkamp, Soeren S. and Brox, Thomas and Ronneberger, Olaf},
  title     = {{3D U-Net}: Learning Dense Volumetric Segmentation from Sparse Annotation},
  booktitle = {Medical Image Computing and Computer-Assisted Intervention (MICCAI)},
  series    = {LNCS},
  volume    = {9901},
  pages     = {424--432},
  publisher = {Springer},
  year      = {2016}
}

@inproceedings{oktay2018attention,
  author    = {Oktay, Ozan and Schlemper, Jo and Le Folgoc, Loic and Lee, Matthew and Heinrich, Mattias and Misawa, Kazunari and Mori, Kensaku and McDonagh, Steven and Hammerla, Nils Y. and Kainz, Bernhard and Glocker, Ben and Rueckert, Daniel},
  title     = {Attention {U-Net}: Learning Where to Look for the Pancreas},
  booktitle = {Medical Imaging with Deep Learning (MIDL)},
  year      = {2018},
  note      = {arXiv:1804.03999}
}

@inproceedings{myronenko2018autoencoder,
  author    = {Myronenko, Andriy},
  title     = {{3D MRI} Brain Tumor Segmentation Using Autoencoder Regularization},
  booktitle = {MICCAI BrainLes Workshop},
  year      = {2018},
  note      = {arXiv:1810.11654}
}

@inproceedings{dosovitskiy2021vit,
  author    = {Dosovitskiy, Alexey and Beyer, Lucas and Kolesnikov, Alexander and Weissenborn, Dirk and Zhai, Xiaohua and Unterthiner, Thomas and Dehghani, Mostafa and Minderer, Matthias and Heigold, Georg and Gelly, Sylvain and Uszkoreit, Jakob and Houlsby, Neil},
  title     = {An Image is Worth 16x16 Words: Transformers for Image Recognition at Scale},
  booktitle = {International Conference on Learning Representations (ICLR)},
  year      = {2021}
}

@inproceedings{tang2022swin,
  author    = {Tang, Yucheng and Yang, Dong and Li, Wenqi and Roth, Holger R. and Landman, Bennett and Xu, Daguang and Nath, Vishwesh and Hatamizadeh, Ali},
  title     = {Self-Supervised Pre-Training of {Swin} Transformers for {3D} Medical Image Analysis},
  booktitle = {IEEE/CVF Conference on Computer Vision and Pattern Recognition (CVPR)},
  pages     = {20730--20740},
  year      = {2022},
  doi       = {10.1109/CVPR52688.2022.02007}
}

@article{cox2024brainsegfounder,
  author  = {Cox, Joseph and Liu, Peng and Stolte, Skylar E. and Yang, Yunchao and Liu, Kang and See, Kyle B. and Ju, Huiwen and Fang, Ruogu},
  title   = {{BrainSegFounder}: Towards {3D} Foundation Models for Neuroimage Segmentation},
  journal = {Medical Image Analysis},
  volume  = {97},
  pages   = {103301},
  year    = {2024},
  doi     = {10.1016/j.media.2024.103301}
}

@article{avants2008syn,
  author  = {Avants, Brian B. and Epstein, Charles L. and Grossman, Murray and Gee, James C.},
  title   = {Symmetric Diffeomorphic Image Registration with Cross-Correlation: Evaluating Automated Labeling of Elderly and Neurodegenerative Brain},
  journal = {Medical Image Analysis},
  volume  = {12},
  number  = {1},
  pages   = {26--41},
  year    = {2008},
  doi     = {10.1016/j.media.2007.06.004}
}

@article{fonov2011unbiased,
  author  = {Fonov, Vladimir and Evans, Alan C. and Botteron, Kelly and Almli, C. Robert and McKinstry, Robert C. and Collins, D. Louis and {Brain Development Cooperative Group}},
  title   = {Unbiased Average Age-Appropriate Atlases for Pediatric Studies},
  journal = {NeuroImage},
  volume  = {54},
  number  = {1},
  year    = {2011},
  doi     = {10.1016/j.neuroimage.2010.07.033}
}

@inproceedings{salehi2017tversky,
  author    = {Salehi, Seyed Sadegh Mohseni and Erdogmus, Deniz and Gholipour, Ali},
  title     = {Tversky Loss Function for Image Segmentation Using {3D} Fully Convolutional Deep Networks},
  booktitle = {Machine Learning in Medical Imaging (MLMI)},
  publisher = {Springer},
  year      = {2017},
  note      = {arXiv:1706.05721}
}

@article{kelly2024radiomic,
  author  = {Kelly, Brendan S. and Mathur, Prateek and McGuinness, Gerard and Dillon, Henry and Lee, Edward H. and Yeom, Kristen W. and Lawlor, Aonghus and Killeen, Ronan P.},
  title   = {A Radiomic ``Warning Sign'' of Progression on Brain {MRI} in Individuals with {MS}},
  journal = {American Journal of Neuroradiology},
  volume  = {45},
  number  = {2},
  pages   = {236--243},
  year    = {2024},
  doi     = {10.3174/ajnr.A8104}
}

@misc{durso2022personalized,
  author       = {Durso-Finley, Joshua and Falet, Jean-Pierre R. and Nichyporuk, Brennan and Arnold, Douglas L. and Arbel, Tal},
  title        = {Personalized Prediction of Future Lesion Activity and Treatment Effect in Multiple Sclerosis from Baseline {MRI}},
  year         = {2022},
  howpublished = {arXiv preprint arXiv:2204.01702}
}

\end{document}